\documentclass[11pt,a4paper]{article}
\usepackage[hyperref]{emnlp-ijcnlp-2019}
\usepackage{times}
\usepackage{latexsym}
\usepackage{graphicx}
\usepackage{multirow}
\usepackage{url}
\usepackage{amssymb}
\usepackage{amsmath}
\usepackage{enumitem}
\usepackage{url}

\aclfinalcopy % Uncomment this line for the final submission
\title{Getting To Know You: User Attribute Extraction from Dialogues}

\author{
    Chien-Sheng Wu, Andrea Madotto, Zhaojiang Lin, Peng Xu, Pascale Fung\\
    Center for Artificial Intelligence Research (CAiRE) \\
    The Hong Kong University of Science and Technology \\
    \texttt{jason.wu@connect.ust.hk} \\}

\date{}

\begin{document}
\maketitle

\begin{abstract}

User attributes provide rich and useful information for user understanding, yet structured and easy-to-use attributes are often sparsely populated. In this paper, we leverage dialogues with conversational agents, which contain strong suggestions of user information, to automatically extract user attributes. Since no existing dataset is available for this purpose, we apply distant supervision to train our proposed two-stage attribute extractor, which surpasses several retrieval and generation baselines on human evaluation. Meanwhile, we discuss potential applications (e.g., personalized recommendation and dialogue systems) of such extracted user attributes, and point out current limitations to cast light on future work.

\end{abstract}

\section{Introduction}

% User attributes are important but hard to get
User attributes are explicit representations of a person's identity and characteristics in a structured format.
They provide a rich repository of personal information for better user understanding in many applications. 
% For example, a recommendation system can deliver products that match what a user may like, and a dialogue system can avoid repeatedly asking similar questions and improve user engagement.
High-quality user attributes are, however, hard to obtain since the information in social networks such as Facebook and Twitter is often sparsely populated~\cite{li2014weakly}. 
Therefore, exploiting unstructured data sources to obtain structured user attributes is a challenging research direction.
% Additionally, better user understanding is a key to build a personalized dialogue agent which has adaptive and proactive capabilities and to improve user engagement as well. 

% Dialogue agent is a good way to collect the data but people never did it
Meanwhile, there is an increasing reliance on dialogue agents to assist, inform, and entertain humans, for example, keeping the elderly company and providing customer service.
Conversational data between users and systems is informative and abundant, and most of the existing deep learning approaches are trained on these large crowd-sourced corpora or scraped conversations.
These models, given the current dialogue context (e.g., few previous turns), are focused on either generating good responses~\cite{serban2015survey}, or incorporating ``system attributes'' to generate consistent responses~\cite{personachat, millionspersona}.
However, the whole dialogue history of the same person is ignored, implying that these systems are not gradually getting to know their users by extracting user information through conversations.

\begin{table}[t]
\resizebox{\linewidth}{!}{%
\begin{tabular}{r|c|c}\hline
 & \textbf{Conversations} & \textbf{User Attributes} \\\hline
\textit{Usr} & Hello, how are you doing today? &  none \\ \hline
\textit{Sys} & I am fine! Where do you live? &  \\ \hline
\textit{Usr} & \begin{tabular}[c]{@{}l@{}}I am originally from California \\ but now I live in Florida for long.\end{tabular} & {(}I, live\_in, Florida{)} \\\hline
\textit{Sys} & \begin{tabular}[c]{@{}l@{}}Florida! You must have a good \\ work-life balance.\end{tabular} &  \\\hline
\textit{Usr} & \begin{tabular}[c]{@{}l@{}}Oh, I no longer work at banks \\ but for exercise I walk often.\end{tabular} & \begin{tabular}[c]{@{}l@{}}{(}I, previous\_profession, banker{)} \\ {(}I, has\_hobby, walking{)}\end{tabular} \\\hline
\textit{Sys} & \begin{tabular}[c]{@{}l@{}}Good to hear that! Do you \\ live with your family?\end{tabular} &  \\ \hline
\textit{Usr} & \begin{tabular}[c]{@{}l@{}}My son. I bring him to church \\ every Sunday with my Ford.\end{tabular} & \begin{tabular}[c]{@{}l@{}}{(}I, has\_children, son{)}\\ {(}I, like\_goto, church{)}\\ {(}I, have\_vehicle, ford{)}\end{tabular} \\\hline
\textit{Sys} & \begin{tabular}[c]{@{}l@{}}Wow sounds good! You \\ can meet many people.\end{tabular} &  \\\hline
\textit{Usr} & \begin{tabular}[c]{@{}l@{}}Sure, but my son is afraid \\ of talking to others.\end{tabular} & {(}My son, misc\_attribute, shy{)} \\
\hline
\end{tabular}
}
\setlength{\belowcaptionskip}{-5pt}
\caption{The conversation column is a daily dialogue between a user and a system. The user attributes column is the potential extracted user information.}
\label{example}
\end{table}

In this paper, we demonstrate that it is feasible to automatically extract user attributes from dialogues.
Given a user utterance, our goal is to predict user information that can be represented as a \textit{(Subject, Predicate, Object)} triplet format, which is available for any downstream application. 
For example, in Table~\ref{example}, \textit{(I, live\_in, Florida)} is extracted from the second user utterance.
Meanwhile, not every utterance has useful information, and some have multiple attributes. 
For instance, ``How are you doing today?'' does not have any user-specific information, but from the fourth user utterance in Table~\ref{example}, we can conclude that the user has a son, likes to go to church, and has a Ford car.
Additionally, unlike standard information extraction tasks, where the extracted information is tagged within the input, some user attributes must be inferred indirectly.
For example, ``My son is afraid of talking to others'' implies that the user's son is a shy person.
 
Since no conversational dataset is available for our purpose, we leverage the state-of-the-art natural language inference (NLI) model to train our model via distant supervision.
Using the existing Persona-Chat dataset~\cite{personachat}, comprising dialogues collected given artificial speaker information called personas, we hypothesize that if an utterance is entailed by a persona sentence, then such a persona sentence can be viewed as a valid user attribute. 
For example, if the persona sentence ``I was a banker'' is entailed by the user utterance ``I no longer work at banks,'' then we can extract the \textit{(I, previous\_profession, banker)} attribute for the utterance.
Although NLI mapping may include some noise, these annotations are cheap and can at least provide a weak source of supervision.

We view user attribute extraction as a pipeline of two tasks: the predicate prediction task and entity generation task. The predicate prediction task first determines whether there is a predicate triggered by a user utterance. 
This is considered as a multi-label classification problem because there could be zero or multiple attributes.
If there is a triggered predicate, then the entity generation task further generates the subject and object phrases to complete the whole user attribute. The subject phrase indicates the ``who'' information, and the object phrase contains the ``what'' information.
We empirically show that our strategy outperforms several retrieval and generation baselines on human evaluation. 
Our contributions are summarized as follows~\footnote{The code is released at \url{ https://github.com/jasonwu0731/GettingToKnowYou}}: 
\begin{itemize}[leftmargin=*]

    \item We are the first to extract user attributes from chit-chat dialogues, which contain strong evidences to suggest users information.

    \item We propose a two-stage attribute extractor that surpasses baselines on human evaluation. We train our model via distant supervision, leveraging an NLI model to obtain cheap and effective training samples.

    \item We discuss potential applications of the extracted user attributes and point out current limitations to cast light on future research directions.

    % \item We propose a new perspective towards lifelong personalized dialogue learning by endowing agents with the ability to extract and accumulate structured user attributes from daily chit-chat conversations. 
    
    % \item A suitable dataset for this task is provided. By pre-training natural language inference (NLI) model using Dialogue Natural Language Inference (DNLI)~\cite{dnli} dataset, we build the mapping between utterances and personas in Persona-Chat~\cite{personachat}.
    
    % \item We propose a pipeline attribute extractor with a copy mechanism that surpasses Seq2Seq, pointer-generator, and key-value retrieval networks. We also demonstrate how to effectively incorporate user attributes into response generation models.
\end{itemize}

% Personalized dialogue agents have received considerable attention since they can make chit-chat more engaging and captivating~\cite{serban2015survey}. 
% There are two perspectives of personalized dialogue agents. 
% The first is giving personalities to dialogue agents ~\cite{personachat,}, and the second, which is rarely discussed, is to adapt to their end users via user attributes. 
% However, high-quality user attributes are hard to obtain~\cite{li2014weakly}. 
% Commonly, the information in social networks is sparsely populated and protected. 
% Therefore, we would like to answer the following question: Can we learn a dialogue agent that can extract rich and structured user information from chit-chat, and accumulate it for better user understanding?
% In this way, the dialogue agent can be more personalized by accumulating the extracted information and utilizing in the future as explicit long-term memory.
% However, the existing approaches are learned to identify speakers or adapt to their fixed, predefined user profiles.

% \begin{figure}[t]
% \centering
% \includegraphics[width=\linewidth]{img/example2.png}
% \caption{An example of extracting structured user attributes from unstructured user utterances.}
% \label{example}
% \end{figure}

\begin{table}[t]
\resizebox{\linewidth}{!}{%
\begin{tabular}{|l|l|}
\hline
\multicolumn{1}{|c|}{Persona A} & \multicolumn{1}{c|}{Persona B} \\ \hline
I just bought a brand new house. & I love to meet new people. \\
I like to dance at the club. & I have a turtle named Timothy. \\
I run a dog obedience school. & My favorite sport is the ultimate frisbee. \\
I have a big sweet tooth. & My parents are living in Bora. \\
I like taking and posting selkies. & Autumn is my favorite season. \\ \hline
\multicolumn{2}{|l|}{\begin{tabular}[c]{@{}l@{}} Conversation \\ {[}A{]} Hi, I just got back from the club.\\ {[}B{]} Cool, this is my favorite time of the year season wise.\\ {[}A{]} I would rather eat chocolate cake during this season.\\ {[}B{]} What club did you go to? Me and Timothy watched TV.\\ {[}A{]} I went to club Chino. What show are you watching?\\ {[}B{]} We watched a show about animals like him.\\ {[}A{]} I love those shows. I am really craving cake.\\ {[}B{]} Why does that matter any? I went outdoors to play frisbee\\ {[}A{]} It matters because I have a sweet tooth.\end{tabular}} \\ \hline
\end{tabular}
}
\setlength{\belowcaptionskip}{-5pt}
\caption{A conversation from the Persona-Chat dataset. Two different personas are provided before they have the conversation below.}
\label{example_persona_chat}
\end{table}

\section{Distant Supervision Data}
\label{sec:dataset}

% Provide one figure about what is persona dataset and the entailment
% and one figure about how the dnli find the mapping and how we use it
% how to determine the predicates

There are no existing dialogue datasets with the labels required for the attribute extraction task. 
Hence, we leverage two datasets, Persona-Chat~\cite{personachat} and Dialogue NLI~\cite{dnli}, to generate distant supervision data. 
We briefly introduce these datasets and discuss some of their limitations.

\paragraph{Persona-Chat}
This is a multi-turn chit-chat corpus with annotation of the participants' personal profiles (e.g., preferences about food, movies).
It is collected by asking two crowd-workers to talk to each other freely but conditioned on their artificial personas, which are established by four to six persona sentences. 
An example from the dataset is provided in Table~\ref{example_persona_chat}.
In total there are 1155 personas with over 5,000 persona sentences, and 162,064 utterances over 10,907 dialogues.
Most of the related works using this dataset~\cite{weston2018retrieve,deepcopy,wolf2019transfertransfo,dinan2019second} focus on adapting systems to a given persona, i.e., learning to generate responses that are consistent with the persona.  

Although the dataset contains pre-defined personas and the corresponding conversations, it cannot be applied directly to the attribute extraction task for the following two reasons:
1) The mapping between utterances and the persona is missing. Which persona sentence is related to which utterance remains unknown.
2) All the personas are written in natural language instead of in a structured format. Natural language description is not easy-to-use for downstream tasks.

\paragraph{Dialogue NLI}
This is a new dataset built upon Persona-Chat~\cite{personachat}, which provides a corpus for NLI task in dialogues. 
The authors demonstrate that consistency of dialogue agents can be improved by re-ranking responses using an NLI model.
Dialogue NLI consists of sentence pairs labeled as entailment, neutral, or contradiction. 
For example, in Table~\ref{example_persona_chat}, the persona sentence ``I like to dance at the club'' for persona A is entailed with the utterance ``I just got back from the club.''

The authors first require human annotation of all the persona sentences in Persona-Chat, mapping into the triplet \textit{($e_1$, $r$, $e_2$)}, where $e_1$ and $e_2$ are entities and $r$ is the relation types. 
They pre-define around 60 different relation types such as \textit{live\_in\_general, like\_food}, and \textit{dislike}.~\footnote{Full relation types are listed in the Appendix} 
For example, the persona sentence ``I just bought a brand new house'' is labeled to the triplet \textit{(I, own, house)}. 
Then they group different persona sentences with the same triplet together. 
Thus sentences in the same group are considered as entailment, and others as neutral and contradiction.

A drawback is that the dataset does not have a human-annotated triplet for each utterance. 
The authors assign a triplet to an utterance by the following criteria: 
1) if its object ($e_2$) is a sub-string of the utterance or 
2) if word embedding similarity between the utterance and the persona sentence is suitably large. 
In this way, they can retrieve a small portion of the utterances that are potentially entailed, but noise is introduced to the dataset and many utterances remain unlabeled.

Since their goal is only to create an NLI dataset, with the strategy mentioned above, the authors are able to collect a large number of training samples. 
On the other hand, our goal is to extract structured attributes from the utterances, and we need as many training samples as possible to learn the mapping. 
Therefore, we need a method to help us find the mapping of the unlabeled utterances. 

\subsection{Combination Strategy}
Our strategy is to combine Persona-Chat and Dialogue NLI. 
We hypothesize that by combining these two datasets, if a user utterance and a persona sentence are positively entailed, then the persona triplet of that persona sentence can be represented as one of the possible user attributes. For example, if the utterance ``I prefer basketball; team sports are fun'' and the persona sentence ``I like playing basketball'' has an entailment relationship, then we assign the triplet of the persona sentence labeled by Dialogue NLI, which is \textit{(I, like\_sports, basketball)}, to be one of the user attributes.

We train an NLI model using the Dialogue NLI corpus, and the trained model can be used as a scorer to predict the entailment score. 
We fine-tune BERT~\cite{devlin2018bert},~\footnote{PyTorch version in \url{github.com/huggingface/pytorch-pretrained-BERT}} a recently proposed pre-trained deep bidirectional Transformer~\cite{transformer}, to predict entailment given two sentences as input.
This scorer achieves 88.43\% test set accuracy on Dialogue NLI, which is aligned (slightly better) with the best-reported model, ESIM~\cite{P17-1152}, with 88.2\% accuracy.

\section{Methodology}
Let us define $N$ utterances in a dialogue as $U = \{u_1,\dots,u_N\}$, where odd and even turns are represented as user utterances and system responses. 
$M$ natural language persona sentences $P = \{p_1,\dots,p_M\}$ in the dataset have their corresponding triplets $T = \{t_1,\dots,t_M\}$. 
Besides persona sentences, each of the utterances may have zero, one or multiple triplets selected from $T$. 
We design a two-stage attribute extractor to obtain \textit{(subject, predicate, object)} triplets from dialogues using a context encoder, a predicate classifier, and an entity generator. 
% The hyper-parameters are reported in the Appendix.

\begin{figure*}[t]
\centering
\includegraphics[width=0.9\linewidth]{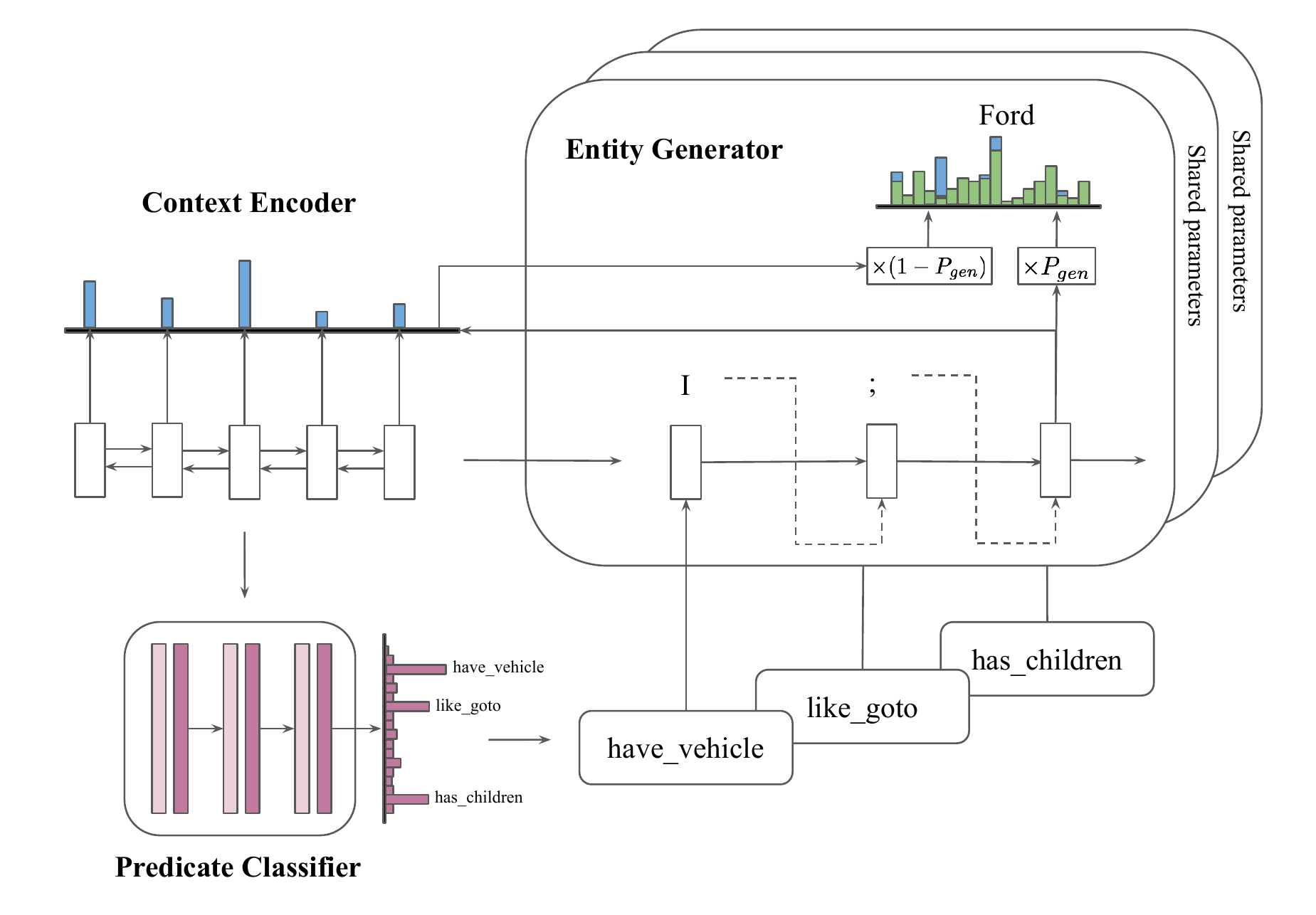}
\caption{The proposed attribute extractor, which has a context encoder, a predicate classifier, and an entity generator. The generator will decode multiple times for every triggered predicate.}
\label{extractor}
\end{figure*}

\subsection{Two-stage Attribute Extractor}
To predict the user attributes, we use a context encoder to capture utterance semantics. 
Then instead of directly generating triplets, we predict all the triggered predicates first. 
Next, an entity generator decodes multiple times for every triggered predicate to obtain their corresponding subject and object phrases. 
For example, in Figure~\ref{example}, three predicates \textit{(have\_vehicle, like\_goto, has\_children)} are triggered by the predicate classifier. Given \textit{have\_vehicle} as input to the entity generator, the subject ``I'' and the object ``Ford'' will be generated.

% We use bi-directional gated recurrent units (GRUs)~\cite{chung2014empirical} as the context encoder to encode a user utterance, and a three-hop end-to-end memory network (MN)~\cite{sukhbaatar2015end} as the predicate classifier. 
% The predicate classifier uses the encoded context as an initial query and all the possible predicates as memory. The last hop memory attention weights are used for the final prediction. Differently, from the original MN, we use a Sigmoid layer on the logits to make a multi-label classification, since each utterance may have multiple predicates triggered. 
% For the entity generator, we use pointer-generator model~\cite{see2017PG} with GRU, which includes a copy mechanism for generating unknown words from input to output.
% For each decoding time step, vocabulary, and source distributions are weighted-summed into one single output distribution by learned scalars. 

\paragraph{Context Encoder}
The context encoder takes a sequence of word embeddings as input and obtains a set of fixed-length vectors $H = (h_1^{enc},\dots,h_l^{enc}) \in \mathbb{R}^{l \times d_{hdd}}$ by bi-directional gated recurrent units (GRUs), where $l$ is the number of words in the utterance and $d_{hdd}$ is the hidden size of the GRU. 
The last hidden state $h_l^{enc}$ is represented as the final encoded vector, which will be used to query the predicate classifier and initialize the entity generator.

\paragraph{Predicate Classifier}
We use a multi-hop ($K=3$ hops) end-to-end memory network (MN)~\cite{sukhbaatar2015end} as our predicate classifier because we believe its reasoning ability can benefit predicates prediction, as shown in question answering and dialogue tasks~\cite{bordes2016learning,dqmem8461426,madotto2018mem2seq,wu2019global}.
We assign the memory in the MN as all the predicate words $R=\{r_1,\dots,r_J\}$, where $J$ is the total number of possible predicates. 
The predicate classifier is queried by the encoded vector $h_l^{enc}$, and the memory attention at each hop $k$ is computed as
% and we take the memory attention weight $\alpha^K$ at last hop as the output distribution as
\begin{equation}
  \alpha^k = Softmax( C^k(P) q^k ) \in \mathbb{R}^{J},
  \label{attn_eq}
\end{equation}
where $C^k$ and $q^k$ are the embedding matrix and query vector at hop $k$, respectively. 
Here, $\alpha^k$ is a soft memory selector that decides the memory relevance with respect to the query vector $q^k$. The model reads out the memory $o^k$ as 
\begin{equation}
  o^k = \sum_i \alpha^k_i C^{k+1}(r_i) \in \mathbb{R}^{d_{hdd}}.
  \label{memread_eq}
\end{equation}
Then the query vector is updated for the next hop using
\begin{equation}
  q^{k+1} = q^k + o^k \in \mathbb{R}^{d_{hdd}}.
  \label{queryupdate_eq}
\end{equation}
In order to perform multi-label classification, instead of taking the \textit{Softmax} function, as in the original MN, to obtain the probability distribution, we replace the \textit{Softmax} layer with a \textit{Sigmoid} layer in Eq.\ref{attn_eq} at the last hop. 
In this way, each of the predicates is triggered separately, and we can predict whether multiple predicate will be triggered, or none of them will be triggered.

\paragraph{Entity Generator}
If a predicate is triggered, our entity generator will generate the corresponding subject and object phrases to complete the final user attribute.
Note that both the subject and object can have more than one word, and we manually concatenate them into one sequence separated by a semicolon. 
For example, we train our model to generate a sequence ``my son; shy'' if the triplet is \textit{(my son, misc\_attribute, shy)}.

Motivated by the multilingual neural machine translation work~\cite{Q17-1024} that uses a single model for all languages but with different start-of-sentence tokens, we also use a single entity generator for all the predicates.
If there are multiple predicates triggered, we decode multiple times using the same parameters for the entity generator with different predicates as input.
In this way, we expect our model to transfer knowledge between different predicate generations.

The first input token of the entity generator is one of the triggered predicates. 
At decoding time step $t$, the generator GRU takes a word embedding $w_t$ as the input and returns a hidden state $h^{dec}_t$. 
The output word distribution $P^{final}_t$ is the weighted-sum of two distributions,
\begin{equation}
  P^{final}_t = P_{gen} P^{vocab}_t + (1-P_{gen}) P^{source}_t,
\label{pg}
\end{equation}
where $P^{vocab}_t=Softmax(W_1 h^{dec}_t)$ is the mapping from the generator hidden states to the vocabulary space using trainable matrix $W_1$, and $P^{source}=Softmax(H h^{dec}_t)$ is the attention weights of the input. The scalar $P_{gen}$ is learned to combine the two distributions,
\begin{equation}
  P_{gen} = Sigmoid(W_2 [h^{dec}_t; w_t; v_c]),
\label{p_gen}
\end{equation}
where $W_2$ is a learned matrix and $v_c= \sum P^{source}_i * h^{enc}_i$ is the context vector.

\subsection{Objective Function}
We use the user attributes obtained from the NLI model as the distant supervision labels.
During training, we optimize the weighted-sum of two loss functions end-to-end, one for the predicate classifier and the other for the entity generator.
The former computes a binary cross-entropy loss $L_p$ between the predicate attention ($\alpha^K$) and the expected ones ($R^{label}$) as
\begin{align}
    \begin{array}{r}
    L_{p} = -\sum_{i} [R^{label}_i \times \log{\alpha^K_i} + \\
    (1-R^{label}_i) \times \log{(1-\alpha^K_i)}]. 
    \end{array}
\end{align}
The latter computes standard cross-entropy loss $L_v$ between the generated sequence ($P^{final}$) and the true subject and object values (defined as $Y^{label}$) as
\begin{equation}
L_v = - \sum_t \log(P^{final}_t (Y^{label}_t)).
\end{equation}
Lastly, we optimize the whole model using the weighted-sum of two losses by a hyper-parameter $\lambda$. The final objective function is
\begin{equation}
Loss = \lambda L_{p} + (1-\lambda) L_{v}.
\end{equation}

\section{Experimental Setup}

\subsection{Training Details}
The attribute extractor is trained using the Adam optimizer~\cite{kingma2014adam} with batch size of 32. 
The learning rate annealing starts from $0.001$ to $0.0001$, and a 0.6 dropout ratio is used. 
All the embeddings are initialized by concatenating Glove embeddings (300)~\cite{pennington2014glove} and character embeddings (100)~\cite{hashimoto2016joint}. 
The $\lambda$ to weight two losses is set to be $0.5$.
A greedy search decoding strategy is used for our entity generator since the generated phrases are usually short.
In addition, to increase model generalization and simulate an out-of-vocabulary setting, a word dropout is applied to the input by randomly masking a small number of input source tokens into unknown tokens.

\subsection{Baselines}
We compare our model with the following implemented baselines: the sequence-to-sequence (Seq2Seq) model~\cite{NIPS2014_5346}, the pointer-generator (PG) model~\cite{see2017PG}, and the key-value memory networks (KVMN)~\cite{miller2016key}. 
Meanwhile, existing OpenIE models, which parse sentences and tag parts of them as output, could be an alternative.
We compare our model with two state-of-the-art open information extraction (OpenIE) pre-trained models, S-OpenIE~\cite{Stanovsky2018SupervisedOI} and LLS-OpenIE ~\cite{P15-1034}.

Seq2Seq, PG, and KVMN are used for internal comparison, where all the models are trained from scratch using the distant supervision data. S-OpenIE and LLS-OpenIE, on the other hand, are used for external comparison, where these two models are trained on several OpenIE datasets and evaluated on the attribute extraction task.
We briefly introduce the baselines:
\begin{itemize}[leftmargin=*]
    \item \textbf{Seq2Seq} is the most common baseline for sequence generation. We use GRUs as a base model to encode a sequence of words and decode a sequence that concatenates \textit{(subject, predicate, object)} by semicolons.
    
    \item \textbf{PG} is one of the best generation models that can copy words from the source text via a pointing mechanism. It computes two distributions (input distribution and vocabulary distribution) and combines them automatically. 
    
    \item \textbf{KVMN} is one of the best neural retrieval models that use memory networks to perform key hashing and value reading. It stores all the pre-defined user attributes in the memory and performs multiple hops before final prediction.
    
    \item \textbf{S-OpenIE} enables a supervised learning approach to the OpenIE task. It formulates OpenIE as a sequence tagging problem. A bi-LSTM transducer and semantic role labeling models are used to extract OpenIE tuples.
    % ~\footnote{\url{https://demo.allennlp.org/open-information-extraction}}
    
    \item \textbf{LLS-OpenIE} first learns a linguistically-motivated classifier to split a sentence into shorter utterances, and produce coherent clauses which are logically entailed by the original sentence. 
    % ~\footnote{\url{https://nlp.stanford.edu/software/openie.html}}
\end{itemize}

\subsection{Evaluation Metrics}
Since we do not have true attributes, even in the test set, we conduct a human evaluation to verify the generated attributes.
Randomly selected utterances from the test set are annotated by three people from Amazon Mechanical Turk.
Turkers are asked to label ``1'' if the attributes can be inferred from the utterance, and otherwise label ``0''. More information about human evaluation is provided in the Appendix.

For reference, we also report the accuracy, F1 score, and BLEU-1 score between the attributes of distant supervision data and the generated attributes.
Accuracy and F1 score are computed by strict matching; i.e., the generated attributes are considered as true positive if and only if every token is exactly the same as the expected attributes. 
The BLEU-1 score~\cite{papineni2002bleu} is, meanwhile, more flexible since the object words do not need to be exactly the same (e.g. ``dogs'' and ``two dogs'', ``dislike heights'' and ``fear of heights''). 

On the other hand, S-OpenIE and LLS-OpenIE are the models pre-trained on other information extraction datasets. 
We conduct a qualitative study with multiple different utterances as input to suggest the fundamental difference in ability between the OpenIE models and ours.

\section{Results}

\begin{table}[t]
\begin{center}
\resizebox{\linewidth}{!}{
\begin{tabular}{r|ccc|c}
 & \textbf{ACC} & \textbf{F1} & \textbf{BLEU-1} & \textbf{Human} \\ \hline
\textit{Seq2Seq} & 7.36 & 21.57 & 41.94 & 31.02 \\ \hline
\textit{PG} & 11.80 & 22.99 & 46.14 & 37.58 \\ \hline
\textit{KVMN} & 25.37 & 27.32 & 40.98 & 52.01 \\ \hline
\textit{Ours} & \textbf{26.52} & \textbf{28.68} & \textbf{51.87} & \textbf{67.11} \\ \hline
\textit{Gold*} & - & - & - &\textit{79.80}
\end{tabular}
}
\end{center}
\caption{Results on user attribute extraction. Our model achieves the highest human evaluation score (statistically significant), outperforming other generation and retrieval models. * Note that the \textit{Gold} row is the distant supervision data.}
\label{UAE-TABLE}
\end{table}

\begin{table}[t]
\begin{center}
\resizebox{0.8\linewidth}{!}{
\begin{tabular}{c|cc}
\multicolumn{1}{r|} {} & \multicolumn{1}{c}{\textbf{ACC}} & \multicolumn{1}{c}{\textbf{F1}} \\ \hline
\textit{Predicate Classifier} & 41.57 & 44.40 \\ \hline
% \textit{PRE} & 0.4598 & 0.4367 \\ \hline
% \textit{REC} & 0.4367 & 0.4586 \\ \hline
\textit{Entity Generator} & 43.48 & 46.03 \\ % \hline
\end{tabular}
}
\end{center}
\setlength{\belowcaptionskip}{-5pt}
\caption{Oracle results of the predicate classifier and entity generator. The entity generator is evaluated given correct predicates as input.}
\label{ORACLE-TABLE}
\end{table}

\subsection{Internal Comparison}
As shown in Table~\ref{UAE-TABLE}, the proposed attribute extraction model achieves the highest F1 score, 28.68\%, which surpasses the other two generation models (Seq2Seq and PG), and it is slightly better than the neural retrieval model (KVMN).
Moreover, our model achieves the highest BLEU-1 score, 51.87, where all the generation models work better than KVMN.
This is because KVMN has the limitation that it can only retrieve triplets that are pre-defined in the dataset, and cannot generate new triplets. 
% And the BLEU-1 score is usually zero if the retrieved attribute of KVMN is not correct.

The oracle study of the attribute extractor is shown in Table~\ref{ORACLE-TABLE}.
The predicate classifier achieves a 44.4\% F1 score on the multi-label classification with 61 possible predicates.
In the oracle study, the entity generator, which is given the correct predicates in the distant supervision data as input, can obtain a 46.03\% F1 score. 
Therefore, the performance drop from 46.03\% to 28.68\% is because of the incorrect predicate prediction.

We also conduct human evaluation over 100 randomly selected test samples.
The results show that 67.11\% of our generated user attributes can be inferred from the user utterances, which is significantly better than KVMN by 15.1\%. 
We also evaluate the distant supervision data, the \textit{Gold} row in Table~\ref{UAE-TABLE}, and the results suggest that around 20\% of the data we use could be noisy input. 

In general, the automatic evaluation scores are not that promising, which suggests that extracting user attributes from dialogue is challenging.
However, since our test data is not human-annotated, these numbers are only for reference.

\begin{table*}[t!]
\begin{center}
\resizebox{\linewidth}{!}{
\begin{tabular}{r|c|c|c}
\hline
 & \textbf{S-OpenIE} & \textbf{LLS-OpenIE} & \textbf{Ours} \\ \hline
Hello, how are you doing tonight? & \textit{(you, doing, tonight)} & \textit{(you, are doing, tonight)} & \textit{none} \\
Yeah, I like cats. I have one. & \textit{(I, have, one)} & \textit{(I, have, one), (I, like, cats)} & \textit{(I, have\_pet, cat)} \\
Go work, so my wife can spend it & \textit{(my wife, spend, it)} & \textit{(my wife, can spend, it)} & \textit{(I, marital\_status, married)} \\
They'd not fit into my mustang convertible & \textit{(my, mustang, convertible)} & \textit{none} & \textit{(I, have\_vehicle, convertible)} \\
I'm originally from California though! & \textit{(I, am, from California)} & \textit{(I, am from, California)} & \textit{(I, place\_origin, California)} \\ \hline
Lol, I like classic cars! & \textit{(lol, like, classic cars)} & \textit{(I, like, cars)} & \textit{(I, like\_music, classic rock)} \\
Tired from too many parties. & \textit{none} & \textit{none} & \textit{(I, like\_activity, partying)} \\
I am well and you? It is cold & \textit{(I, am, well), (it, is, cold)} & \textit{(it, is, cold)} & \textit{(I, like\_general, cold weather)} \\

I traveled a lot, I even studied abroad. & \textit{(I, travel, a lot), (I, even studied, aboard)} & \textit{none} & \textit{none} \\ \hline

\end{tabular}
}
\end{center}
\setlength{\belowcaptionskip}{-5pt}
\caption{External comparison of our attribute extractor and two existing open information extraction approaches, S-OpenIE and LLS-OpenIE. Both positive and negative examples are provided.}
\label{SAMPLE-TABLE}
\end{table*}

\subsection{External Comparison}
We show some generated samples from the test set in Table~\ref{SAMPLE-TABLE}, and compare them with  S-OpenIE~\cite{Stanovsky2018SupervisedOI} and LLS-OpenIE ~\cite{P15-1034} to suggest the difference. 
One can observe that existing OpenIE approaches directly parse words from sentences, but our model learns to predict possible predicates.
For example, our model successfully predicts \textit{none} if none of the predicates is triggered, but others still return the parsing results, which contain important information. 
In addition, our model is able to predict relations which are not explicitly mentioned in the sentences. 
For example, the user utterance ``I like cats. I have one'' triggers the predicate \textit{have\_pet}, and ``My wife can spend it'' triggers the predicate \textit{marital\_status}.

We also provide some negative examples of our generated user attributes.
We find three common errors: wrong predicate prediction, ambiguous attribute inference, and missing attribute prediction.
First, if our model does not predict predicates correctly, it may generate out-of-context object phrases.
For example, our model predicts \textit{like\_music} as a triggered predicate for the utterance ``I like classic cars!'' because it is biased by people mentioning classical music.
Second, we find that in some cases our model generates attributes that are relevent but not certain, making the attribute ambiguous.
For example, when a user says he/she is ``Tired from too many parties,'' our model predicts the attribute \textit{(I, like\_activity, partying)} although the user does not mention it explicitly.
Third, sometimes no predicate is triggered, even if there is some useful user information.
For example, we should be able to conclude that a user likes to travel if he/she says ``I travel a lot. I even studied abroad.''

\section{Discussion}
Once we obtain user attributes, they can be applied to many downstream applications, for example, search, friend recommendation, online advertisement, computational social science, personalized personal assistant, etc.
We select two directions we are interested in and discuss them in detail, and point out current limitations.

\subsection{Potential Applications}

\paragraph{Personalized Dialogue Agents}
These systems have received considerable attention since they can make chit-chat more engaging and captivating~\cite{serban2015survey}. 
There are two perspectives on personalized dialogue agents: the first is giving personalities to the agents~\cite{personachat, millionspersona}, and the second, which is rarely discussed, is to adapt the agents to their end users via user attributes. Therefore, if we can endow a dialogue system with a user attribute extraction module, we can make a step towards lifelong personalized dialogue systems.

A dialogue system can view user attributes extracted from the history as explicit long-term memory.
This information is able to avoid the system repeating the same or similar questions.
For example, if a user mentioned ``I was born in September 2009'' in a previous conversation two days ago, a personalized dialogue system should avoid asking similar questions, such as ``Which month is your birthday?'' and ``How old are you?'' 
In addition, such attributes can be used to filter or suggest what the system should reply.
For example, it would not be appropriate for a personalized system to ask ``How is your university life?'' if the user was born in 2009 and it is 2019. 
It would be better for the system to reply ``Wow! Soon you will be ten years old!'' after inferring the time information.

\paragraph{Personalized Recommender System}
% Over the past few years, recommender systems have been more and more effective in helping people identify their preferred resources from a large number of candidate objects.
% To build such systems, you need a dataset of items and users and ideally also interactions of users with items.
There are three main common systems for personal recommendation: 
% knowledge-based, content-based, and collaborative filtering.
A knowledge-based system has both user and item attributes, and make recommendations based on user-item attribute similarities;
A content-based system recommends items similar to those a given user has liked in the past, regardless of the preferences of other users;
A collaborative filtering system, meanwhile, is based on past interactions of the whole user-base, e.g., examining k-nearest neighbor users.

Most of these recommender systems require real online interactions of users with items, such as mouse clicking and browsing.
Our approach provides an alternative way to collect user attributes ``offline,'' which can then be applied to cluster users, or record items that a user has mentioned in the past. 
For example, if both users are from San Francisco and they all like baseball, we can recommend a Giants game to one user if the other mentions it often.

\subsection{Current Limitations}
% only 61 predicates
We have presented the idea of extracting user attributes from daily dialogues. 
Although our two-stage model with distant supervision can achieve reasonable results, we believe there exist limitations that should be addressed in the future. 
% We summarize them in the following three points.

Most importantly, a suitable dialogue dataset with clean attribute extraction labels is needed. 
First of all, using the NLI model to determine the relation mapping between persona sentences and utterances is not an ideal solution. As we mentioned in the error analysis, there is an ambiguous attribute inference problem. This problem suggests that using the entailment model may not always capture the real causality information. For example, the fact that a person attends many parties does not necessarily mean they like parties.
Next, the pre-defined predicates from~\citet{dnli} are not collected comprehensively, which may not be able to cover all the relations in a real scenario. Therefore, using clustering techniques to group  more predicates automatically is an appealing solution.
Lastly, the conversations in the Persona-Chat dataset are not collected naturally, with most of the users tending to ignore what the other said and just talking about themselves. Therefore, it is hard to evaluate whether ``understanding your partner'' helps agents speak properly. 
Also, since there is no publicly available data with the same user continually talking to a system, it is hard to evaluate the lifelong setting. 

\section{Related Work}

\paragraph{User Attributes Inference}
Most previous work has treated user attribute inference from social media as a classification task, such as gender prediction~\cite{ciot2013gender}, age prediction~\cite{rao2010classifying,alekseev2016predicting}, occupation~\cite{preoctiuc2015analysis}, and political polarity~\cite{pennacchiotti2011machine, johnson-goldwasser-2016-know}.
\citet{li2014weakly} propose to extract three user attributes (spouse, education, and job) from Twitter using weak supervision.
\citet{bastian2014linkedin} present a large-scale topic extraction pipeline, which includes constructing a folksonomy of skills and expertise on LinkedIn.

\paragraph{Information Extraction}
% \noindent\textbf{Information Extraction}:
Closed and open form information extraction are important and well studied NLP tasks~\cite{Banko, P10-1013, P11-1062, Fader}. 
Both rule-based~\cite{D12-1048, DelCorro} and learning-based~\cite{C14-1220,D15-1206,P15-1034,P16-1123,Stanovsky2018SupervisedOI,D18-1157} methods have been proposed by the research community.
However, most approaches are only able to handle information by tagging/parsing part of the input source. 
Additionally, our work is also related to the dialogue state tracking tasks for task-oriented dialogue systems~\cite{WuTradeDST2019}.
% To the best of our knowledge, we are the first to tackle information extraction using the generative model in chit-chat system.

\paragraph{Personalized Systems}
% \noindent\textbf{Dialogue Systems}:
Recommender systems predict the preference a user would give to an item, which is utilized in a variety of areas.
Content-based filtering~\cite{pazzani2007content}, knowledge-based filtering~\cite{burke2000knowledge} and collaborative filtering~\cite{sarwar1998using} are the most common approaches for recommender systems.
For dialogue applications, \citet{lucas2009managing} and \citet{joshi2017personalization} focus on letting the agent be aware of the human pre-defined profile and so adjust the dialogue accordingly. 
\citet{zemlyanskiy2018aiming} define a mutual information discovery score to re-rank system generating responses.
\citet{MadottoPAML} uses meta-learning to fast adapt to unseen persona scenarios.

% \citet{li2016persona} learn speaker embeddings for
% speaker consistency and \citet{personachat} propose the Persona-chat dataset for chit-chat systems. Several approaches are explored based on this dataset ~\cite{kulikov2018importance,yavuzdeepcopy,hancock2019learning}, including ConvAI2 challenge~\cite{dinan2019second}. 
% , however, very few of them addressed the problem of attribute extraction from the dialogue. 

\section{Conclusion}
We utilize conversational data to extract user attributes for better user understanding. Due to lacking a labeled dataset, we apply distant supervision with a natural language inference model to train our proposed two-stage attribute extractor. Our model surpasses several retrieval and generation baselines on human evaluation, and is different from existing open information extraction approaches. In the end, we discuss potential downstream applications and point out current limitations to provide suggestions for future work. 

% We propose to make a step towards lifelong personalized dialogue agents, by learning to extract user attributes from chit-chat and accumulate them for better user understanding.
% A new dataset is created for this task based on Persona-Chat and DNLI datasets.
% The proposed pipeline attribute extractor outperforms several existing baselines in terms of automatic and human evaluation. 
% We also show the effectiveness of incorporating extracted attributes to support response generation.
% Future work will be to manually collect high-quality attribute labels for the large-scale conversational dataset.

% \section*{Acknowledgments}
% The acknowledgments should go immediately before the references.  Do
% not number the acknowledgments section. Do not include this section
% when submitting your paper for review. \\

\bibliography{acl2019}
\bibliographystyle{acl_natbib}

\end{document}